\documentclass[a4paper]{article}

\usepackage{INTERSPEECH2021}
\pdfoutput=1
\usepackage{xcolor}
\usepackage{soul} 

\usepackage[noadjust]{cite}

\title{Evaluation of Audio-Visual Alignments in Visually Grounded Speech Models}
\name{Khazar Khorrami$^1$, Okko Räsänen$^{1,2}$}

\address{
  $^1$Unit of Computing Sciences, Tampere University, Finland\\
  $^2$Department of Signal Processing and Acoustics, Aalto University, Finland}
\email{khazar.khorrami@tuni.fi, okko.rasanen@tuni.fi}

\begin{document}

\maketitle
 
\begin{abstract}
Systems that can find correspondences between multiple modalities, such as between speech and images, have great potential to solve different recognition and data analysis tasks in an unsupervised manner. This work studies multimodal learning in the context of visually grounded speech (VGS) models, and focuses on their recently demonstrated capability to extract spatiotemporal alignments between spoken words and the corresponding visual objects without ever been explicitly trained for object localization or word recognition. 
As the main contributions, we formalize the alignment problem in terms of an audiovisual alignment tensor that is based on earlier VGS work, introduce systematic metrics for evaluating model performance in aligning visual objects and spoken words, and propose a new VGS model variant for the alignment task utilizing cross-modal attention layer. We test our model and a previously proposed model in the alignment task using SPEECH-COCO captions coupled with MSCOCO images. We compare the alignment performance using our proposed evaluation metrics to the semantic retrieval task commonly used to evaluate VGS models. We show that cross-modal attention layer not only helps the model to achieve higher semantic cross-modal retrieval performance, but also leads to  substantial improvements in the alignment performance between image object and spoken words.  

\end{abstract}

\section{Introduction}

Utilization of statistical dependencies between different modalities has the potential to replace or supplement supervised learning in many tasks, as the data streams can potentially be used as weak supervision for each other.
As an example of such multimodal systems, so-called visually grounded speech (VGS) models have been recently proposed  [\citen{harwath2016unsupervised,harwath2017learning,harwath2018jointly,chrupala2017representations,merkx2019language, mortazavi2020speech}].
They are unsupervised audiovisual
algorithms that can learn shared semantic concepts between visual and speech data, 
given a series of unlabeled images and utterances describing them (e.g., \cite{harwath2016unsupervised,chrupala2017representations}). As a result, the models can be used to search for semantically similar content in audio and images. VGS models are also interesting for modeling of infant language learning, which is essentially an unsupervised multimodal learning process with learners having access to both auditory and visual input  (see, e.g., \cite{rasanen2019computational,khorrami2021can}).

Although VGS models typically operate based on high-dimensional semantic embeddings from which spatial and temporal characteristics of the data have been abstracted away, the architecture of these models can also be modified to estimate alignments between visual objects and the corresponding spoken words in the input data. For instance, \cite{harwath2018jointly} introduced a latent audiovisual tensor that assigns explicit alignment scores on each spatial position and time-step for input pairs of images and utterances.  One could also attempt to extract the object-word-alignments from other latent (e.g., audiovisual attention) layers of the models. This makes VGS models promising for unsupervised decoding (segmentation) of image and audio data into their constituent units in an unsupervised manner, and opens up a largely unexplored strategy to learn structural properties of initially unorganized and unlabeled data (e.g., visual object categorization or unsupervised language learning). 

Localization of objects within images using text-based queries has been investigated in different lines of research, such as in object detection given a category label \cite{redmon2017yolo9000} and in natural language object retrieval \cite{karpathy2014deep, qiao2020referring,hu2016natural,luo2017comprehension,rohrbach2016grounding,mao2016generation}. Given a query word, these models either directly try to predict the locations of object bounding boxes in images \cite{hu2016natural,rohrbach2016grounding}, or by ranking alternative image regions in terms of the potential presence of the relevant object \cite{mao2016generation}. Even though the majority of existing multi-modal works have focused on phrase-based object detection, the multimodal learning and alignment task can also be viewed as symmetric with respect to both input modalities, i.e., learned objects in the visual input could also be used to segment correct words in the other modality, and vice versa. 
Moreover, object localization has been mainly evaluated using intersection over union (IoU) between the predicted and true bounding boxes of visual objects. IoU uses hard decisions on what pixels are detected, potentially followed by another thresholding to determine if a specific object is detected or not (see, e.g., \cite{mao2016generation}). 

In \cite{harwath2018jointly}, the proposed VGS model was evaluated using a number of heuristic approaches, including measuring the capability of the audiovisual tensor to conduct speech-based object localization, to form meaningful audio-visual pattern clusters, and in building an object-word concept dictionary. Although functional, these evaluation strategies are variable, dependent on the characteristics of the selected small test set, and require manual analysis of the results. Hence, in order to enable systematic comparison of alternative models and facilitate further unsupervised multimodal alignment system development, standardized definition of the alignment problem and associated evaluation metrics would be desirable.

Given this background, our present goal is to formally define the problem of multimodal alignment between  speech and images, here studied in the context of VGS models, and to define evaluation metrics accord with the problem definition. We start from audiovisual alignment tensors (derived from \cite{harwath2018jointly}), and define two metrics that capture complementary aspects of the alignment performance in terms of finding visual objects for spoken words (or phrases), and finding words in speech, given visual objects. We also propose a new VGS model variant for unsupervised audiovisual alignment using audiovisual attention, and show that it outperforms the models from \cite{harwath2018jointly} in both alignment and semantic retrieval tasks. Although we place our work in the context of VGS models, the proposed methodology is applicable to both explicit and implicit alignment models, and should generalize beyond image and speech modalities. %
\section{Problem and evaluation definition}

\begin{figure}[t]
  \centering
  \includegraphics[scale=0.35]{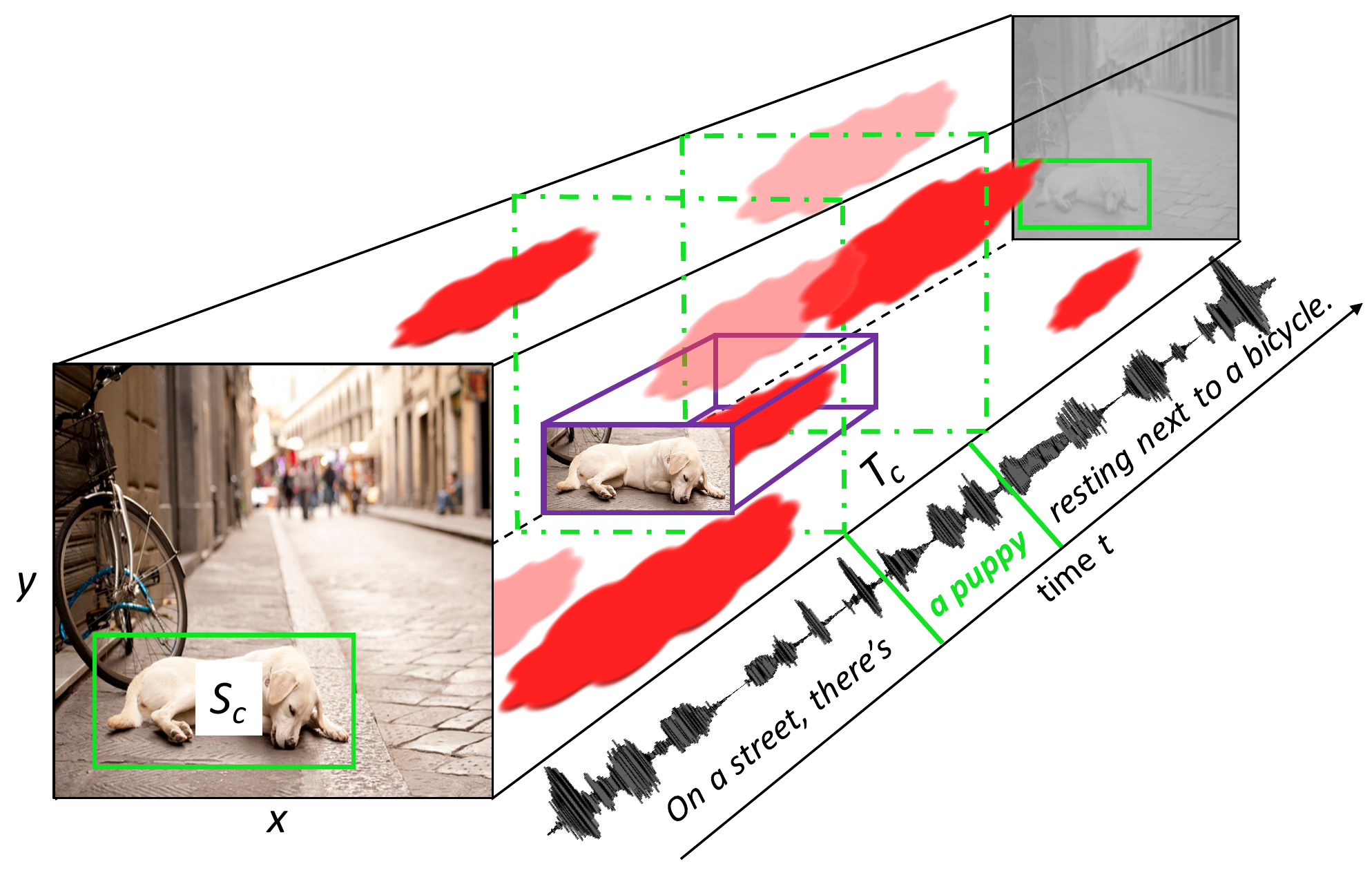}
  \vspace{-12pt}
  \caption{An illustration of a temporospatial alignment tensor \textbf{T}[x ,y, t] between an image and an utterance related to the image. Red "clouds" illustrate manifolds where T has positive values. Ground-truth alignment region for [dog] is visualized with a violet box. Conceptually adapted from \cite{harwath2018jointly}.}
  \label{figtensor}
  
  \vspace{-12 pt}
  
\end{figure}

The basic goal of the alignment process is to identify the link between spoken words or phrases and the corresponding visual objects or concepts in images. Formally, we define this link using an audiovisual alignment tensor \begin{math}\mathbf{T}[x,y,t] \in \begingroup\mathbb{R}\endgroup_{\geq0}^3, x \in \{1,...,N_x\}, y \in \{1,...,N_y\}, t \in \{1,...,T\}, \end{math} which defines the association strength between each pixel \{\textit{x,\,y}\} in the image space with each time step \textit{t} in the speech signal (see Fig. \ref{figtensor} for illustration). \begin{math}N_x \times N_y\end{math} is the size of the image in pixels, and \textit{T} is the duration of the utterance in terms of 10-ms signal frames.  In \begin{math}\mathbf{T}[x,y,t]\end{math}, zeros stand for no association and the larger the (positive) value, the stronger the model's confidence that the pixels and timestamps are related to each other. The goal of an alignment model is to derive this tensor, given an input image and an utterance. In practice, the tensor may be a final output of a model (e.g., using supervised training) or extracted as a latent representation of a VGS model (as in \cite{harwath2018jointly} and all models in our present experiments).

Evaluation of the alignment is based on ground-truth-knowledge on visual objects that are related to specific acoustic words or phrases, basically word/phrase timestamps together with object pixel masks. Given a concept ("class") \textit{c}, we define \{\textit{x},\,\textit{y}\}\begin{math}\in S_c\end{math} as the set of pixels corresponding to visual object \begin{math}o_c\end{math}, and \begin{math}\{t_\textup{onset}, ..., t_\textup{offset}\}\in T_c\end{math} as the set of time-frames corresponding to the related word \begin{math}w_c\end{math}. Together they define a subset of \begin{math}\mathbf{T}[x,y,t]\end{math} corresponding to the ground-truth alignment. 

We propose two new primary metrics to evaluate how well \begin{math}\mathbf{T}\end{math} relates to the ground-truth: \textit{alignment score (AS)} and \textit{glancing score (GS)}, both measured separately in time and space.

\subsection{Alignment score}

AS measures how well the model attends to the correct visual object(s) throughout a spoken word, or how well the attention across all the pixels of an object focuses on the time-instances of the corresponding word. In the case of object alignment given a spoken utterance, \begin{math}
AS_{\textup{object}}(c) \in [0,1]\end{math} is calculated as follows.

First, each frame in \begin{math}\mathbf{T}\end{math} is normalized to sum up to one
\begin{math}
\mathbf{T}'[x,y,t]=\mathbf{T}[x,y,t]/\sum_{x,y}^{}{\mathbf{T}[x,y,t]}
\end{math} to measure the distribution of "attention" across the image at each time step. Then the proportion of attention on the target object is obtained by

\vspace{-8pt}

\begin{equation}
AS_{\textup{object}}(c) = \frac{1}{|T_c|}\sum_{S_c\,,T_c}^{}\mathbf{T}'[x,y,t]
\end{equation}

\vspace{-4pt}

\noindent where \begin{math}{|T_c|}\end{math} is the number of frames in \begin{math}w_c\end{math}. As a result, \begin{math}
AS_{\textup{object}}(c)\end{math} obtains a value of 0 if none of the attention is on the target object \begin{math}o_c\end{math} during the word \begin{math}w_c\end{math}, and 1 if all attention is located on the object \begin{math}o_c\end{math} during the entire duration of \begin{math}w_c\end{math}.

Word alignment score \begin{math}
AS_{\textup{word}}(c) \in [0,1]\end{math} is calculated in an analogous manner, now first normalizing each pixel of \begin{math}\mathbf{T}\end{math} to sum up to one across the utterance: \begin{math}
\mathbf{T}''[x,y,t]=\mathbf{T}[x,y,t]/\sum_{t}^{}\mathbf{T}[x,y,t]
\end{math}, and then measuring the average temporal overlap of the resulting scores and \begin{math}w_c\end{math} across the \begin{math}o_c\end{math} pixels:

\vspace{-8pt}

\begin{equation}
AS_{\textup{word}}(c) = \frac{1}{|S_c|}\sum_{S_c,T_c}^{}\mathbf{T''}[x,y,t]
\end{equation}

\vspace{-4pt}

If \begin{math}
AS_{\textup{word}}(c)\end{math} is 0, the temporal attention at all the pixels of \begin{math}o_c\end{math} have zero value during \begin{math}w_c\end{math}, whereas score of 1 means that attention at \textit{all} of the \begin{math}o_c\end{math} pixels is always completely within the bounds of \begin{math}w_c\end{math} and zero during other time steps.

\subsection{Glancing score}

The alignment score enforces each word time-step or object pixel to have consistent attention on the corresponding target in the other signal modality. However, in some use cases we may only be interested whether the model "takes a look" at the correct object once it recognizes a word, or tags the correct word in time given a partial observation of an object. As an example, a temporally causal model simulating human eye-gaze during speech comprehension could "glance" at the correct object once it recognizes the corresponding word \cite{Golinkoff_2013}. However, such a model would not be able to maintain sustained attention on the target object before identifying the word as it unfolds in time. To evaluate such a behavior, we introduce glancing score GS.

In object glancing score \begin{math}GS_{\textup{object}} \in [0,1]\end{math}, the cumulative attention map across the image during the entire word \begin{math}w_c\end{math} is first calculated as \begin{math}
\textbf{A}_c[x,y]=\sum_{t\in T_c}^{}\mathbf{T}[x,y,t] 
\end{math}. 
This is then normalized to a spatial distribution proper \begin{math}\textbf{A}'_c[x,y]=\textbf{A}_c[x,y]/\sum {_{x,y}\textbf{A}_c[x,y]}\end{math}, and compared to ground-truth pixels to measure the proportion of attention on the target:
\vspace{-8pt}
\begin{equation}
GS_{\textup{object}}(c)=\sum_{\left \{ x,y \right \} \in S_c}^{}\textbf{A}'_c[x,y]
\end{equation}

\vspace{-8pt}

\noindent The corresponding word glancing score \begin{math}GS_{\textup{word}} \in [0,1]\end{math} is obtained by first measuring the total object attention as a function of time by \begin{math}
a[t]=\sum_{\left \{ x,y \right \} \in S_c
}^{}\mathbf{T}[x,y,t]
\end{math} and normalizing it to have a sum of 1 across the utterance with \begin{math}a'[t]=a[t]/\sum_{t} a[t]\end{math}. The score is then obtained by

\vspace{-8pt}

\begin{equation}
GS_{\textup{word}}(c)=\sum_{t \in T_c  } a'_c[t] .
\end{equation}

\vspace{-4pt}

\noindent In essence, \begin{math}GS_{\textup{object}} \end{math} is the proportion of attention within the spatial extent of object \begin{math}o_c\end{math} during word \begin{math}w_c\end{math} whenever \begin{math}\mathbf{T} > 0\end{math}, but the attention does not have to be defined (positive-valued) throughout the duration of the word. In an analogous manner, \begin{math}GS_{\textup{word}} \end{math} corresponds to the proportion of attention on the target word \begin{math}w_c\end{math} compared to other words, given the pixels of \begin{math}o_c\end{math}, but not all pixels have to have positive attention values in \textbf{T}. The more there is relative attention outside the correct targets, the lower the scores will be. If \begin{math}\mathbf{A}[x,y]\end{math} or \begin{math}a[t]\end{math} fully zero before normalization, they are replaced by a uniform distribution across their elements. 

Equations above describe scoring for individual word-object pairs \begin{math}w_c\end{math}-\begin{math}o_c\end{math} in individual images-utterance samples. In practice, the score should be calculated for each ground-truth pair in each image-utterance-pair of the test set. Also note that confusion errors for both AS and GS can be derived from the equations simply by comparing \begin{math}S_{c_1}\end{math} and \begin{math}T_{c_2}\end{math}, where \begin{math}c_1 \neq  c_2\end{math}.

\section{Model Description}

Our models follow the main structure of VGS models (see, e.g., \cite{harwath2017learning, harwath2018jointly, chrupala2017representations}), where two branches of neural layers embed data from speech and image domains, respectively. Then the modality-specific vector embeddings are mapped into a shared semantic space. Input to the system consists of images paired with spoken descriptions of them. Output is a similarity score indicating the semantic relatedness of the input pairs. The network is trained using a triplet loss \cite{harwath2016unsupervised} that tries to assign higher scores to semantically related image-speech pairs compared to unrelated pairs, and with the maximum separation limited by a margin \textit{M}. 

However, unlike other common VGS models (e.g., \cite{harwath2016unsupervised,chrupala2017representations,merkx2019language}), encoder outputs of the so-called DAVEnet model \cite{harwath2018jointly} maintain modality-specific signal representations as a function of input speech time (1-D) and image position (2-D). The embedding vectors are then mapped ("aligned") together through matrix product, resulting a 3-D tensor \textbf{T}[\textit{w}, \textit{h}, \textit{n}] spanning in both spatial location (\textit{w}, \textit{h}) and time frames (\textit{n}). This audio-visual alignment tensor \textbf{T} allows the model to co-localize patterns within both modalities. The overall similarity score between input speech and images can then be obtained by taking the sum and/or maximum of \textbf{T} in time and/or spatial dimensions,
resulting in three different model criteria for training: MISA (max over image, sum over audio), SIMA (sum over image, max over audio), SISA (sum over image, sum over audio) \cite{harwath2018jointly}, whereas maxing over both dimensions appears to be difficult to train.

In our experiments, the first model variant is similar to the DAVEnet (here:  \textit{CNN${_\textup{0}}$}).  
In CNN${_\textup{0}}$, the speech encoder is stack of five convolutional layers (with layer sizes of [128, 256, 256, 512, 512]) with gradual temporal downsampling with maxpooling over time. VGG16 model \cite{russakovsky2015imagenet}, up to the last convolutional layer and pretrained on ImageNet data, is used as the image encoder. This is followed by one trainable 2D-convolution layer with 512 filters. Both the speech and image encoder branches are then fed to a dense linear layer, followed by L2-normalization, and then combined into \textbf{T} with matrix product.

\begin{figure}[t]
  \centering
  \includegraphics[width=\linewidth]{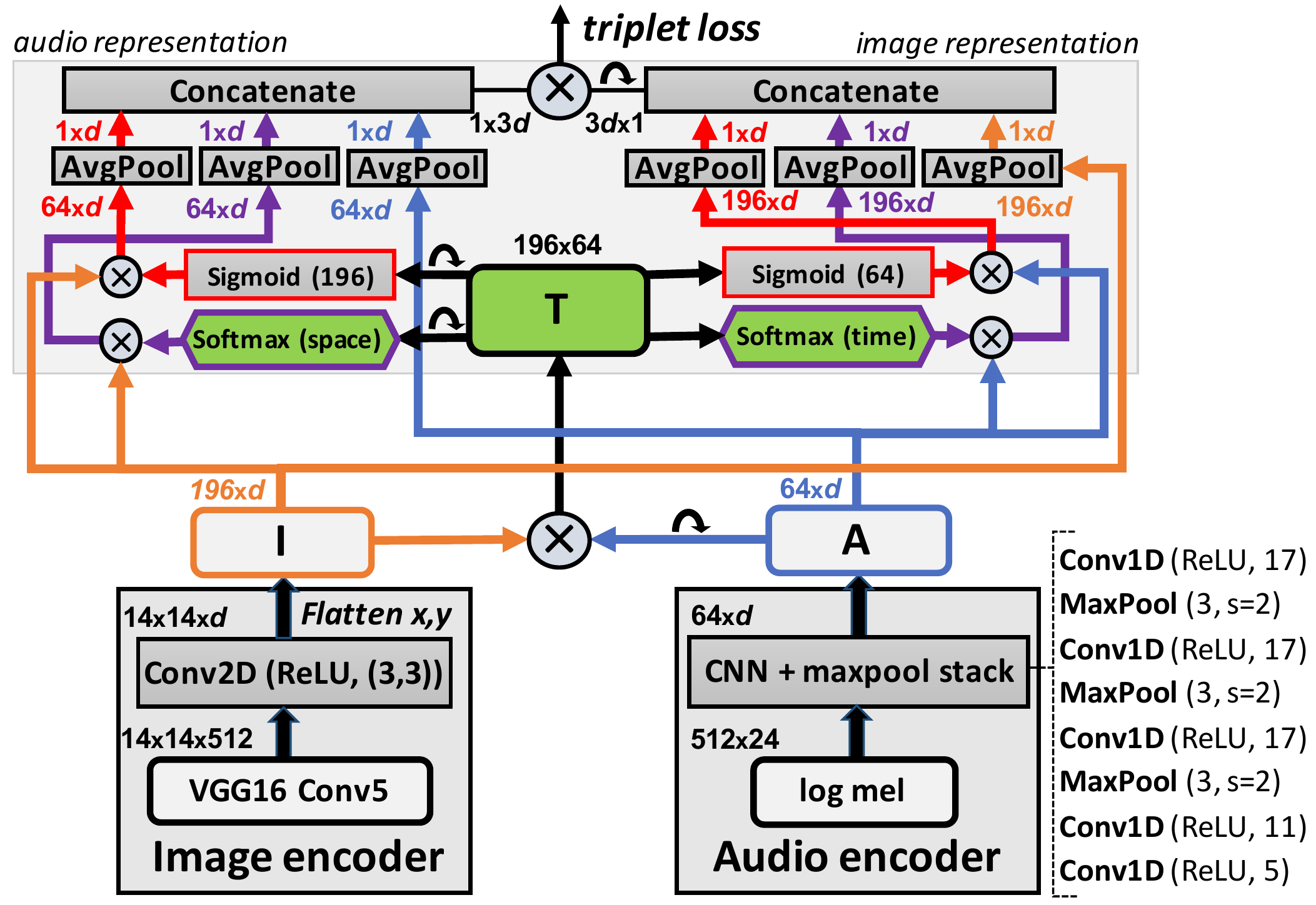}
  \caption{CNN${_\textup{ATT}}\textup{v0}$ architecture using a cross-modal attention block on top of DAVEnet model. 64 = number of time frames, 196 = flattened spatial coordinates (x,y), $\curvearrowright$ = transpose. The layers used in alignment evaluation are highlighted with green. 
  }
  \vspace{-12pt}
  \label{fig:model.arch}
\end{figure}

Our CNN${_\textup{ATT}}$  (Fig. \ref{fig:model.arch}) is obtained from CNN${_\textup{0}}$ by adding a cross-modal attention layer on top of the \textbf{T}[\textit{w}, \textit{h}, \textit{n}]
to help the model to attend to specific image objects and spoken words using the information from the another modality. This is inspired by the fact that the audiovisual tensor \textbf{T} introduced in \cite{harwath2018jointly} is similar to scoring function applied in dot product attention mechanism \cite{luong2015effective,vaswani2017attention}. Thus, we extended the model to have a complete attentional module by applying softmax non-linearity separately in space (for a query in time; Fig. \ref{fig:model.arch} left branch) and in time (for a query in space; Fig. \ref{fig:model.arch} right branch) in order to produce a distribution of weights over image space and speech frames, respectively. In parallel to the softmax, we also apply a dense layer with a sigmoid activation function to produce an extra attentional representation, as we found this to outperform the use of softmax only in the retireval task. These attention scores are then used to produce corresponding spatially weighted representations for audio (left) and time-weighted representations for image (right), followed by average pooling to get rid of absolute positional information. In the final stage, outputs of the softmax and sigmoid layer attentions are concatenated with the original image and speech representations to produce the final speech and image embeddings. These embeddings are then L2-normalized and compared using a similarity score (dot product) in triplet-loss training, and alignments can be extracted from \textbf{T} or after taking the softmax in time or space.

We tested three alternative variations for the CNN${_\textup{ATT}}$ architecture: CNN${_\textup{ATT}}$v0 with the same speech encoder as the CNN${_\textup{0}}$, CNN${_\textup{ATT}}$v1 with the first maxpool layer removed to obtain \textbf{T} at a higher temporal resolution of 128 frames (compared to 64 frames in other variants), and CNN${_\textup{ATT}}$v2 which was equal to CNN${_\textup{ATT}}$v0 but with less filters ([64, 128, 256, 256, 512]). 

\section{Experimental setup}

For our experiments, we used MSCOCO \cite{lin2014microsoft}, which consists of images of everyday objects and their contexts, together with their speech-synthesized captions from SPEECH-COCO \cite{havard2017speech}. 
The dataset includes a total of 123,287 images with 91 common object categories (e.g. dog, pizza, chair) from 11 super-categories (e.g., animal, food). In SPEECH-COCO, each image is paired by five synthetic speech captions describing the scene using the object categories. In our experiments, we used the training set of MSCOCO images and their verbal descriptions for model training and validation. 
The original validation set of $\sim$40k images was used as a held-out test set for evaluation, using one randomly chosen spoken caption per image. 

SPEECH-COCO contains metadata on words and their timestamps, while MSCOCO contains manually annotated pixel masks for its 80 visual object categories. However, there is no direct one-to-one mapping between image object labels (e.g., [dog]) and spoken words (e.g., "puppy"). We derived ground-truth pairings of objects and words automatically using semantic similarities from a pre-trained Word2vec model \cite{mikolov2013efficient}. We first extracted nouns of the caption transcripts using NLTK-toolbox, and then used Word2vec cosine similarity between words and object labels as a means of identifying semantically matching pairs. Words and objects with Word2vec similarity above a threshold of 0.5 were considered as a ground-truth word-object pair \begin{math}w_c\end{math}-\begin{math}o_c\end{math} (a concept). Selection of this threshold was based on manual observation of the similarity histograms across the dataset, where 0.5 provided a reasonable cutoff point in the somewhat bimodal (but noisy) similarity distribution. 

Model training followed the same triplet-loss protocol as in [\citen{harwath2016unsupervised,harwath2017learning,harwath2018jointly,chrupala2017representations,khorrami2021can}].
Adam optimizer was used for both models initial learning rate of $lr=$10e-4 and triplet loss margin $M=0.1$. In the speech processing channel, we applied ReLU activation after each convolutional layer, followed by batch normalization.  We  measured \textit{recall@10} audiovisual semantic retrieval score \cite{hodosh2013framing} using the representation similarity scores to ensure that the models have learned the semantic relationships between the two modalities. For each variant, we saved the best model based on the recall score of the validation set. 
In both CNN${_\textup{0}}$ and CNN${_\textup{ATT}}$ models, recall scores were measured using speech and image embedding layers before audiovisual \textbf{T} (I and A in Fig. \ref{fig:model.arch}). AS and GS scores were then measured for each word-object pair in the test set. Total score was calculated as the average of the 80 class-specific scores, where each class-based score was computed as the average score of all samples belonging to that class. We also report baseline results for a system that assigns random values to \textbf{T} (from uniform [0, 1]). 

For the CNN${_\textup{0}}$ and CNN${_\textup{ATT}}$, AS and GS scores were measured for the 
tensor \textbf{T}[\textit{w},\textit{h},\textit{t}]. For CNN${_\textup{ATT}}$, we also measure alignments from the \textbf{T} after the attention module softmax functions (Fig. \ref{fig:model.arch}, green boxes), applied in spatial and temporal directions for the object and word detection tasks, respectively. For both models, we upsampled \textbf{T}[\textit{w},\textit{h},\textit{n}] to \textbf{T}[\textit{x},\textit{y},\textit{t}] to match with the original image size and the target 10-ms time resolution of the evaluation protocol.

\section{Results}

\begin{figure}[t]
  \centering
  \includegraphics[width=\linewidth]{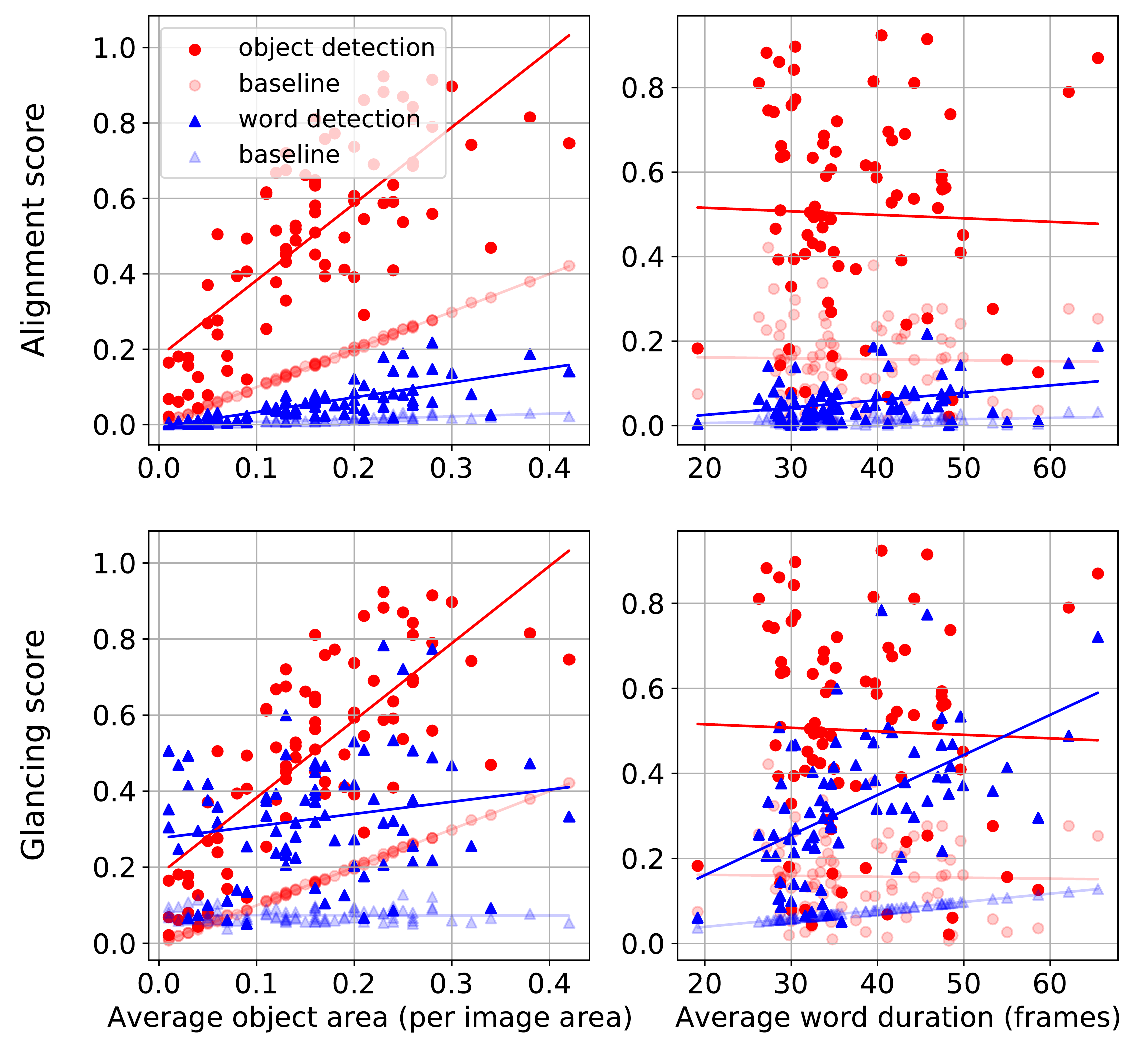}
  \vspace{-24 pt}
  \caption{Relationship of alignment score (top) and glancing score (down) with average object size and average word duration (for CNN${_\textup{ATT}}$v1 softmax). Solid lines = linear fits to the data. Shaded lines = corresponding fits to random baselines.}
  \label{fig:scatter}
  \vspace{-12 pt}
\end{figure}

Table \ref{tab:recall} shows the recall@10 scores 
obtained for the different models for both speech-to-image and image-to-speech search. 
The scores show that all the models learned to solve the audiovisual retrieval task, and that the performance is higher for CNN${_\textup{ATT}}$ than CNN${_\textup{0}}$. Results for the CNN${_\textup{ATT}}$v2 also show that the use of cross-modal attention leads to higher retrieval performance with a smaller number of parameters than in CNN${_\textup{0}}$. In general, CNN${_\textup{ATT}}$ models were faster to train and converged before 70 epochs. In comparison, a maximum of 150 epochs was required for the CNN${_\textup{0}}$ to converge.

Table \ref{tab:scores} shows the results
for the object and word alignment and glancing scores together with the random baseline. As can be seen, audiovisual tensors obtained from all model variants can produce alignment and glancing scores clearly above what is obtained by chance. However, all the scores are higher for CNN${_\textup{ATT}}$.  
Comparing the CNN${_\textup{0}}$ variants, one can also observe that maxpooling across time or space (SIMA and MISA) improves from the SISA variant in the respective dimension. In fact, we also tried to train MIMA (maxpooling in time and space simultaneously), but the model did not train well (in terms of validation recall@10) even with a large number of epochs.

In terms of CNN${_\textup{ATT}}$, aligning and glancing scores increase substantially for the softmax weighted tensor compared to the preceding audiovisual tensor. In fact, among all tested variants, only the softmax tensors produced alignment and glancing scores above 0.5. This further indicates that application of cross-modal attention helps the model to focus on semantically relevant image objects and spoken words, and that the model finds the correspondences between two modalities more accurately than without attention. As can be seen, both aligning and glancing scores increase for the CNN${_\textup{ATT}}$v1 model (compared to CNN${_\textup{ATT}}$v0) with increased temporal resolution of \textbf{T}. 

\begin{table}[th]
  \caption{Recall@10 scores for speech-to-image  and image-to-speech search tasks for the compared models. "T-dim" refers to the number of time frames in \textbf{T} for encoding 5.12 s of speech.}
  \vspace{-8 pt}
  \label{tab:recall}
  \centering
  \resizebox{7.7cm}{!}{
  \begin{tabular}{l c c c c }
   
    {\textbf{Model}} & {\textbf{ parameters}} & {\textbf{T-dim}} &
    {\textbf{\shortstack{speech-to \\ image}}} &
    {\textbf{\shortstack{image-to \\ speech}}} \\
    
  \hline
  CNN${_\textup{0}}$ &  11,072,128 &  64 &  -   & - \\
  \hline
  \hspace{3mm}SISA  & & & 0.489 & 0.487 \\
  \hspace{3mm}MISA  & & & 0.436  & 0.305 \\
  \hspace{3mm}SIMA  & & & 0.401 & 0.446 \\
   
  \hline
  CNN${_\textup{ATT}}$ v0  & 10,587,540 & 64 & \textbf {0.562} & \textbf{ 0.559}\\
  CNN${_\textup{ATT}}$ v1  & 13,943,508 & 128 & 0.547 & 0.531\\
  CNN${_\textup{ATT}}$ v2  & 6,401,748 & 64 & 0.536 & 0.544\\
  \end{tabular}}
  \end{table}
  
 \vspace{-12 pt}

\begin{table}[th]
  \caption{Alignment and glancing scores for the model variants.}
  \vspace{-8 pt}
  \label{tab:scores}
  \centering
  \resizebox{7.2cm}{!}{
  \begin{tabular}{ l | r r r r }
   
    {\textbf{Model}} &
    
    {$AS_{\textup{object}}$} &
    {$AS_{\textup{word}}$} &
    {$GS_{\textup{object}}$} &
    {$GS_{\textup{word}}$} \\
  \hline
  baseline   & 0.158 & 0.011 & 0.158 & 0.073 \\ 
  \hline

 CNN${_\textup{0}}$ & & & &  \\
    \hline
   \hspace{3mm}SISA    & 0.200 & 0.016 & 0.200 & 0.102 \\
   \hspace{3mm}MISA    & 0.270 & 0.020 & 0.267 & 0.132 \\
   \hspace{3mm}SIMA    & 0.223 & 0.041 &0.222 & 0.214\\
     \hline 
    CNN${_\textup{ATT}}$ &  & & &  \\
   \hline
  \hspace{3mm}v0   & 0.285 & 0.028 &0.282 & 0.168 \\ 
  
  \hspace{3mm}v1   & 0.292 & 0.038 & 0.295 & 0.227 \\  
   
  \hspace{3mm}v2  & 0.279 & 0.030 & 0.277 & 0.180 \\  
  
  \hspace{3mm}v0 softmax   & 0.504 & 0.052 & 0.504 & 0.317 \\
  \hspace{3mm}v1 softmax  & \textbf{0.518} & \textbf{0.076}  & \textbf{0.518} & \textbf{0.446} \\ 
     \hspace{3mm}v2 softmax & 0.501 & 0.056 & 0.501 & 0.327 \\

  \end{tabular}
  }
\end{table}

Finally, Fig. \ref{fig:scatter} illustrates how the AS and GS depend on ground truth visual object size and word duration (for CNN${_\textup{ATT}}$v1 softmax). Both object detection (red) and word detection (blue) scores increase with increasing visual object size. They also increase more than what is expected by chance with larger targets. Fig. \ref{fig:scatter} also shows that AS and GS for word detection increase for longer words, whereas the object detection scores are not affected by the word duration. Similar patterns were  observed across all the compared models. 

\section{Conclusions}

In this paper, we proposed alignment and glancing scores as two novel metrics for evaluating performance of visually grounded speech (VGS) systems in an alignment task between visual objects and spoken words. We also introduced a VGS model variant based on cross-modal attention, and compared how different VGS models perform on the alignment task using our proposed metrics. The results show that the metrics can capture different aspects of cross-modal alignments, and that the attention-based VGS models outperform the earlier non-attentional alignment approach in both alignment and semantic retrieval tasks.  

\section{Acknowledgements}
This research was funded by Academy of Finland grants no. 314602 and 320053.

\bibliographystyle{IEEEtran}

\bibliography{main}

\begin{thebibliography}{10}
\providecommand{\url}[1]{#1}
\csname url@samestyle\endcsname
\providecommand{\newblock}{\relax}
\providecommand{\bibinfo}[2]{#2}
\providecommand{\BIBentrySTDinterwordspacing}{\spaceskip=0pt\relax}
\providecommand{\BIBentryALTinterwordstretchfactor}{4}
\providecommand{\BIBentryALTinterwordspacing}{\spaceskip=\fontdimen2\font plus
\BIBentryALTinterwordstretchfactor\fontdimen3\font minus
  \fontdimen4\font\relax}
\providecommand{\BIBforeignlanguage}[2]{{%
\expandafter\ifx\csname l@#1\endcsname\relax
\typeout{** WARNING: IEEEtran.bst: No hyphenation pattern has been}%
\typeout{** loaded for the language `#1'. Using the pattern for}%
\typeout{** the default language instead.}%
\else
\language=\csname l@#1\endcsname
\fi
#2}}
\providecommand{\BIBdecl}{\relax}
\BIBdecl

\bibitem{harwath2016unsupervised}
D.~F. Harwath, A.~Torralba, and J.~R. Glass, ``Unsupervised learning of spoken
  language with visual context,'' in \emph{{Proceedings of the Advances in
  Neural Information Processing Systems 29 (NIPS 2016)}}, December 5--10,
  Barcelona, Spain, 2016, pp. 1858--1866.

\bibitem{harwath2017learning}
D.~Harwath and J.~R. Glass, ``Learning word-like units from joint audio-visual
  analysis,'' in \emph{{Proceedings of the 55th Annual Meeting of the
  Association for Computational Linguistics ({ACL} 2017)}}, July 30--August 4,
  Vancouver, Canada, 2017, pp. 506--517.

\bibitem{harwath2018jointly}
D.~Harwath, A.~Recasens, D.~Sur{\'{\i}}s, G.~Chuang, A.~Torralba, and J.~R.
  Glass, ``Jointly discovering visual objects and spoken words from raw sensory
  input,'' in \emph{{15th European Conference on Computer Vision (ECCV 2018)}},
  September 8--14, Munich, Germany, 2018, pp. 659--677.

\bibitem{chrupala2017representations}
G.~Chrupa{\l}a, L.~Gelderloos, and A.~Alishahi, ``Representations of language
  in a model of visually grounded speech signal,'' in \emph{{Proceedings of the
  55th Annual Meeting of the Association for Computational Linguistics (Volume
  1: Long Papers)}}, July 30--August 4, Vancover, Canada, 2017, pp. 613--622.

\bibitem{merkx2019language}
D.~Merkx, S.~L. Frank, and M.~Ernestus, ``Language learning using speech to
  image retrieval,'' in \emph{{Proceedings of the 20th Annual Conference of the
  International Speech Communication Association (Interspeech 2019)}},
  September 15--19, Graz, Austria, 2019, pp. 1841--1845.

\bibitem{mortazavi2020speech}
M.~S. Mortazavi, ``Speech-image semantic alignment does not depend on any prior
  classification tasks,'' in \emph{Proceedings of the 21th Annual Conference of
  the International Speech Communication Association (Interspeech 2020)},
  October 25--29, Shanghai, China, 2020, pp. 3515--3519.

\bibitem{rasanen2019computational}
O.~R{\"{a}}s{\"{a}}nen and K.~Khorrami, ``A computational model of early
  language acquisition from audiovisual experiences of young infants,'' in
  \emph{{Proceedings of 20th Annual Conference of the International Speech
  Communication Association (Interspeech 2019)}}, Graz, Austria, September
  2019, pp. 3594--1598.

\bibitem{khorrami2021can}
\BIBentryALTinterwordspacing
K.~Khorrami and O.~R{\"a}s{\"a}nen, ``Can phones, syllables, and words emerge
  as side-products of cross-situational audiovisual learning?-a computational
  investigation,'' \emph{submitted for publication}. [Online]. Available:
  \url{https://psyarxiv.com/37zna}
\BIBentrySTDinterwordspacing

\bibitem{redmon2017yolo9000}
J.~Redmon and A.~Farhadi, ``Yolo9000: better, faster, stronger,'' in
  \emph{Proceedings of the IEEE conference on computer vision and pattern
  recognition}, July 21--26, Honolulu, Hawaii, 2017, pp. 7263--7271.

\bibitem{karpathy2014deep}
A.~Karpathy, A.~Joulin, and L.~Fei-Fei, ``Deep fragment embeddings for
  bidirectional image sentence mapping,'' in \emph{Proceedings of the 27th
  International Conference on Neural Information Processing Systems -- Volume
  2}, 2014, pp. 1889--1897.

\bibitem{qiao2020referring}
Y.~Qiao, C.~Deng, and Q.~Wu, ``Referring expression comprehension: A survey of
  methods and datasets,'' \emph{IEEE Transactions on Multimedia}, online early
  access, 2020.

\bibitem{hu2016natural}
R.~Hu, H.~Xu, M.~Rohrbach, J.~Feng, K.~Saenko, and T.~Darrell, ``Natural
  language object retrieval,'' in \emph{Proceedings of the IEEE Conference on
  Computer Vision and Pattern Recognition}, June 27--30, Las Vegas, Nevada,
  2016, pp. 4555--4565.

\bibitem{luo2017comprehension}
R.~Luo and G.~Shakhnarovich, ``Comprehension-guided referring expressions,'' in
  \emph{Proceedings of the IEEE Conference on Computer Vision and Pattern
  Recognition}, July 21--26, Honolulu, Hawaii, 2017, pp. 7102--7111.

\bibitem{rohrbach2016grounding}
A.~Rohrbach, M.~Rohrbach, R.~Hu, T.~Darrell, and B.~Schiele, ``Grounding of
  textual phrases in images by reconstruction,'' in \emph{European Conference
  on Computer Vision}, October 11--14, Amsterdam, Netherlands, 2016, pp.
  817--834.

\bibitem{mao2016generation}
J.~Mao, J.~Huang, A.~Toshev, O.~Camburu, A.~L. Yuille, and K.~Murphy,
  ``Generation and comprehension of unambiguous object descriptions,'' in
  \emph{Proceedings of the IEEE Conference on Computer Vision and Pattern
  Recognition}, June 27--30, Las Vegas, Nevada, 2016, pp. 11--20.

\bibitem{Golinkoff_2013}
R.~M. Golinkoff, W.~Ma, L.~Song, and K.~Hirsh-Pasek, ``Twenty-five years using
  the intermodal preferential looking paradigm to study language acquisition:
  what have we learned?'' \emph{Perspectives on Psychological Science}, vol.~8,
  pp. 316--339, 2013.

\bibitem{russakovsky2015imagenet}
O.~Russakovsky, J.~Deng, H.~Su, J.~Krause, S.~Satheesh, S.~Ma, Z.~Huang,
  A.~Karpathy, A.~Khosla, M.~S. Bernstein, A.~C. Berg, and F.~Li, ``Imagenet
  large scale visual recognition challenge,'' \emph{International Journal of
  Computer Vision}, vol. 115, pp. 211--252, 2015.

\bibitem{luong2015effective}
M.-T. Luong, H.~Pham, and C.~D. Manning, ``Effective approaches to
  attention-based neural machine translation,'' in \emph{Proceedings of the
  2015 Conference on Empirical Methods in Natural Language Processing},
  September 17--21, Lisbon, Portugal, 2015, pp. 1412--1421.

\bibitem{vaswani2017attention}
A.~Vaswani, N.~Shazeer, N.~Parmar, J.~Uszkoreit, L.~Jones, A.~N. Gomez,
  L.~Kaiser, and I.~Polosukhin, ``Attention is all you need,'' in \emph{31st
  Conference on Neural Information Processing Systems (NIPS 2017)}, December
  4--9, Long Beach, California, 2017.

\bibitem{lin2014microsoft}
T.-Y. Lin, M.~Maire, S.~Belongie, J.~Hays, P.~Perona, D.~Ramanan,
  P.~Doll{\'a}r, and C.~L. Zitnick, ``{Microsoft COCO: Common objects in
  context},'' in \emph{{European Conference on Computer Vision (ECVV 2014)}},
  September 6--12, Zurich, Switzerland, 2014, pp. 740--755.

\bibitem{havard2017speech}
\BIBentryALTinterwordspacing
W.~Havard, L.~Besacier, and O.~Rosec, ``{SPEECH-COCO:} 600k visually grounded
  spoken captions aligned to {MSCOCO} data set,'' \emph{arXiv pre-print}, 2017.
  [Online]. Available: \url{http://arxiv.org/abs/1707.08435}
\BIBentrySTDinterwordspacing

\bibitem{mikolov2013efficient}
T.~Mikolov, K.~Chen, G.~Corrado, and J.~Dean, ``Efficient estimation of word
  representations in vector space,'' in \emph{{Proceedings of 1st International
  Conference on Learning Representations ({ICLR} 2013)}}, May 2--4, Scottsdale,
  Arizona, 2013.

\bibitem{hodosh2013framing}
M.~Hodosh, P.~Young, and J.~Hockenmaier, ``Framing image description as a
  ranking task: Data, models and evaluation metrics,'' \emph{Journal of
  Artificial Intelligence Research}, vol.~47, pp. 853--899, 2013.

\end{thebibliography}

\end{document}